\pdfoutput=1

\documentclass[11pt]{article}

\usepackage[preprint]{acl}

\usepackage{times}
\usepackage{hyperref}
\usepackage{latexsym}

\usepackage[T1]{fontenc}

\usepackage[utf8]{inputenc}

\usepackage{microtype}

\usepackage{inconsolata}

\usepackage{graphicx}

\usepackage{url}
\usepackage[size=tiny, disable]{todonotes}
\usepackage{subcaption}
\usepackage{booktabs}
\usepackage{makecell}
\usepackage{stfloats}
\usepackage{float}
\usepackage{enumitem}
\newcommand*{\kgcarbon}[1]{#1~kg~CO$_2$-eq}

\usepackage{bm}
\usepackage{amsfonts}
\newcommand\scalemath[2]{\scalebox{#1}{\mbox{\ensuremath{\displaystyle #2}}}}

\usepackage{amsmath}

\DeclareMathOperator*{\argmin}{arg\,min}

\newcommand{\se}[1]{\textcolor{black}{#1}}
\newcommand{\senew}[1]{\textcolor{black}{#1}}

\newcommand{\dl}[1]{\textcolor{black}{#1}}
\newcommand{\dln}[1]{\textcolor{black}{#1}}

\title{xCOMET-lite: Bridging the Gap Between Efficiency and Quality in Learned MT Evaluation Metrics}

\author{
Daniil Larionov\textsuperscript{1} \quad  
Mikhail Seleznyov\textsuperscript{4,3} \quad 
Vasiliy Viskov\textsuperscript{3} \quad \\ 
\textbf{Alexander Panchenko\textsuperscript{3,4}} \quad 
\textbf{Steffen Eger\textsuperscript{1,2}}\\
\textsuperscript{1}~NLLG, University of Mannheim, \textsuperscript{2} University of Technology Nuremberg, \textsuperscript{3}~Skoltech, \textsuperscript{4}~AIRI \\
\href{mailto:daniil.larionov@uni-bielefeld.de}{daniil.larionov@uni-mannheim.de}
}

\begin{document}
\maketitle
\begin{abstract}
State-of-the-art trainable machine translation evaluation metrics like xCOMET achieve high correlation with human judgment but rely on large encoders (up to 10.7B parameters), making them computationally expensive and inaccessible to researchers with limited resources. To address this issue, we investigate whether the knowledge stored in these large encoders can be compressed while maintaining \senew{quality}. We employ distillation, quantization, and pruning techniques to create efficient xCOMET alternatives and introduce a novel data collection pipeline for efficient black-box distillation. Our experiments show that, using quantization, xCOMET can be compressed up to three times with \dln{no} quality degradation. Additionally, through distillation, \dln{we create an 278M-sized xCOMET-lite metric, which has only 2.6\% of xCOMET-XXL parameters, but retains 92.1\% of its quality. Besides, it surpasses strong small-scale metrics like COMET-22 and BLEURT-20 on \senew{the} WMT22 metrics challenge dataset \senew{by 6.4\%}, despite using 50\% \senew{fewer} parameters.} All code, dataset, and models are \href{https://github.com/NL2G/xCOMET-lite}{available online}.

\end{abstract}

\setlength{\belowdisplayskip}{2pt} \setlength{\belowdisplayshortskip}{2pt}
\setlength{\abovedisplayskip}{2pt} \setlength{\abovedisplayshortskip}{2pt}

\section{Introduction}
\par Automatic evaluation metrics are crucial for reliably measuring the quality of responses from natural language generation (NLG) systems. Researchers and practitioners working on tasks such as machine translation (MT), summarization, poetry generation, \se{etc.}, routinely use metrics to assess their systems' quality. Apart from directly assessing the systems, evaluation metrics have many other applications:
{\bf a)} filtering web-scale datasets~\citep{peter-etal-2023-theres}; {\bf b)} using metric\se{s} as reward function\se{s} for Reinforcement Learning~\citep{DBLP:journals/corr/abs-2401-08417}; {\bf c)} \se{o}nline re-ranking of outputs of multiple systems to choose the best response to return to the user~\citep{fernandes-etal-2022-quality}.
\begin{figure}[h]
    \centering
    \setlength{\belowcaptionskip}{-15pt}
    \includegraphics[width=\columnwidth]{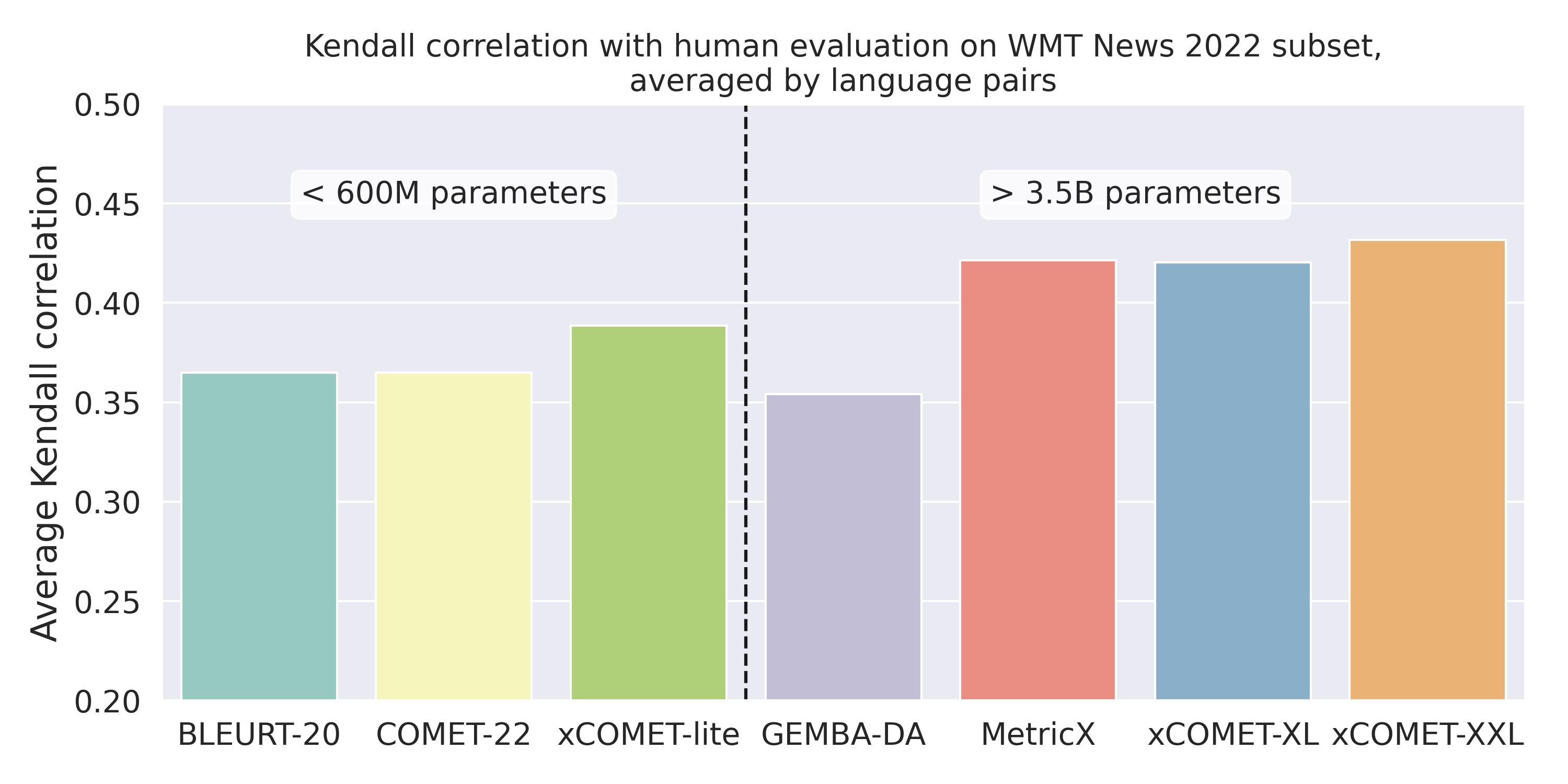}
    \caption{xCOMET can be distilled into a small model, which will be 6-7 percentage points better than SOTA models with comparable parameter count.}
    \label{fig:distillation}
\end{figure}
\par With generative models' growing sizes and complexity, automatic evaluation metrics also evolve and become more computationally expensive. \se{In the last few years,} for MT evaluation, researchers \se{have moved} from traditional n-gram and character-based metrics, such as BLEU~\citep{DBLP:conf/acl/PapineniRWZ02} and chrF~\citep{popovic-2015-chrf}, to embedding-based metrics, such as BERTScore~\citep{DBLP:conf/iclr/ZhangKWWA20} and MoverScore~\citep{zhao-etal-2019-moverscore}, to learned metrics, which provide state-of-the-art correlation with human judgment. According to~\citet{freitag-etal-2023-results}, the best-performing metrics for MT evaluation are xCOMET~\citep{DBLP:journals/corr/abs-2310-10482}, MetricX~\citep{juraska-etal-2023-metricx}, and GEMBA-MQM~\citep{kocmi-federmann-2023-gemba}. All those metric models have a large number of parameters: xCOMET and MetricX have 10.7B-13B parameters, while GEMBA-MQM relies on \se{the} Large Language Model (LLM) GPT4~\citep{DBLP:journals/corr/abs-2303-08774}, for which the number of parameters is unknown but speculated to be around 1.7T\footnote{\url{https://twitter.com/soumithchintala/status/1671267150101721090}}.
\par The lack of efficient alternatives to these models creates a disparity in access among researchers. Under-resourced labs, students, startups, and hobbyists without access to top-tier accelerators (with more than 22GB VRAM) or financial resources for paid API\se{s} cannot employ those metrics. Those with access to such resources may also experience prolonged iteration time due to the computation needed for those models. 
\se{This} 
is especially noticeable in the case of repeated evaluations during the hyperparameter optimization or processing of large-scale datasets. For instance, running the xCOMET-XXL model to filter a crawled dataset of $10^7$ examples would take 142.2 hours on a capable consumer-grade GPU, requiring 42.6 kWh of electricity and emitting around \kgcarbon{15.6}.\footnote{Assumptions: GPU power draw of 350W, 0.05s per example on average and \kgcarbon{0.368}/kWh US power grid carbon intensity taken as reference\se{.}} Thus, developing alternative efficient metrics is now more vital than ever.

\par In this paper, we explore various techniques to develop an efficient alternative to the state-of-the-art xCOMET metric for evaluating MT quality. Our approach focuses on three main methods: knowledge distillation, quantization, and pruning. Knowledge distillation is a method of creating capable small deep fitted models by training them on the outputs of the larger model. We apply knowledge distillation~\citep{DBLP:journals/corr/HintonVD15}, training a smaller version of the xCOMET model on large amounts of data, using labels 
created by the original xCOMET-XXL model. Quantization reduces the precision of deep learning model parameters and activations from 32/16 bits into 8, 4, 3, and 2 bits, occupying less memory and allowing for faster computations. Pruning involves the removal of less significant parts of the model, either specific parameters, blocks of parameters, or entire layers. We apply layer pruning together with subsequent fine-tuning, which allows for accelerated inference throughput and helps mitigate potential accuracy loss. By exploring distillation, quantization, and pruning, as well as their combinations, we aim to create an efficient alternative to xCOMET that maintains a high level of quality while substantially reducing hardware requirements.

\par Our main contributions are as follows: {\bf a)} we conduct a comprehensive study of different compression methods (knowledge distillation, quantization, and pruning) and their interactions for the state-of-the-art MT evaluation metric xCOMET. To the best of our knowledge, this is the first work to systematically investigate the effectiveness and trade-offs of these techniques when applied to a large-scale, complex metric like xCOMET; {\bf b)} we introduce a novel data collection pipeline for preparing large-scale, high-quality datasets for black-box distillation of xCOMET. We collect 14M examples with translation hypotheses of varying quality paired with high-quality reference translations. This enables the distilled model to effectively transfer the evaluation capabilities of the teacher model, xCOMET-XXL;
{\bf c)} through our distillation method, we develop xCOMET-lite, a lightweight yet highly effective MT evaluation metric. xCOMET-lite achieves state-of-the-art quality among metrics with < 600M parameters, surpassing the previous best model, COMET-22, while being \senew{substantially} 
smaller;
{\bf d)} we explore the use of quantization for compressing xCOMET and demonstrate that 3-bit quantization can effectively reduce hardware requirements for 3B and 11B model versions without compromising quality;
{\bf e)} we investigate the effectiveness of pruning for compressing xCOMET and show that while pruning up to 25\% of the model layers can improve inference speed and memory consumption with only a marginal impact on quality, removing more layers leads to substantial quality degradation. {\bf f)} \senew{W}e conduct a novel study of the interactions between compression methods, revealing that distillation combines well with quantization but is incompatible with pruning in our experiments.

\section{Related Work}

Recent work has explored improving the transparency and capabilities of MT evaluation metrics.~\citet{juraska-etal-2023-metricx} introduced MetricX. This learned regression-based metric achieves state-of-the-art correlations with human judgments through multi-stage fine-tuning on direct assessment data, consolidated MQM scores, and small-scale synthetic corpora, which is used to boost robustness. It is based on the mT5-XXL encoder-decoder model with 11B parameters.~\citet{kocmi-federmann-2023-gemba} proposed GEMBA-MQM, which leverages the GPT-4 language model with a few-shot prompting approach to identify translation error spans and categories. 

This enables detailed error analysis, though reliance on the computationally expensive proprietary GPT-4 LLM poses challenges for academic research.~\citet{DBLP:journals/corr/abs-2310-10482} developed xCOMET, a learned metric based on the XLM-RoBERTa-XL/XXL encoder that bridges sentence-level quality prediction with fine-grained error span detection. By training on direct assessment and MQM data, xCOMET achieves top quality on sentence-level, system-level, and error span prediction tasks while providing interpretability through its predicted error spans.

Previously, researchers have also explored techniques for creating more {\bf efficient} MT evaluation metrics while preserving their correlation with human judgments. \citet{kamal-eddine-etal-2022-frugalscore} proposed FrugalScore, which learns lightweight versions of metrics like BERTScore and MoverScore using knowledge distillation. Their distilled metrics perform similarly to the originals while being much faster and having orders of magnitude fewer parameters. \citet{rei-etal-2022-searching} introduced COMETINHO, a more compact and faster version of the COMET metric. They optimize the COMET code using caching and length batching and further compress the model using pruning and knowledge distillation on synthetic data. The resulting model is 80\% smaller and over 2 times faster than the original while maintaining competitive quality. \dl{These works studied knowledge distillation and pruning in application to rather simple and small-scale regression metrics like BERTScore, MoverScore, and COMET.}

In \senew{contrast}, we scale and adapt these methods for \se{the} much larger state-of-the-art xCOMET metric, which generates error annotations in addition to the quality scores. \senew{Further,} we apply a more comprehensive pruning approach than \senew{the} one applied in~\citet{rei-etal-2022-searching} as\senew{,} after pruning layers\senew{,} we do a fine-tuning of a small selection of model parameters, which substantially reduces performance drop. Additionally, we are \senew{the} first to study quantization \senew{for} MT evaluation metrics, as well as do an interaction analysis between compression methods.

\citet{larionov-etal-2023-effeval} provide a comprehensive evaluation of techniques for improving the efficiency of MT evaluation metrics. They explore replacing computation-heavy transformers with more lightweight variants, employing linear and quadratic approximations for alignment algorithms, and using adapters for parameter-efficient training of the COMET metric. Their experiments across multiple datasets and metrics demonstrate that distilled models like TinyBERT can provide an optimal balance of quality and efficiency, with the parallel adapter configuration yielding the best results for the COMET metric training. Unlike this paper, we study compression methods that help boost inference-time efficiency for learned metrics, which helps reduce compute costs and environmental footprint on a much broader scale.

\section{Method\se{s}}
\se{We explore} three compression techniques to develop an efficient alternative to \senew{xCOMET} for evaluating \se{MT} quality: quantization, pruning, and distillation. These methods aim to reduce the computational requirements and improve the inference speed of xCOMET while maintaining a high level of \se{quality}. 

\paragraph{Quantization}\dl{Quantization is a highly effective compression method with two main approaches: quantization-aware training (QAT) and post-training quantization (PTQ)~\citep{DBLP:journals/corr/abs-2106-08295}. QAT offers better prediction quality but requires costly training, making PTQ more popular. PTQ is further divided into data-free and data-aware methods, where the latter relies on calibration to estimate \se{the} data distribution parameters for higher prediction quality. Another distinction is weight-only quantization and weight \& activation quantization, with the second approach having slightly lower prediction quality but potential for faster computations using efficient 8- or 4-bit CUDA kernels.}

In a nutshell, the quantization process comes down to finding bias and scale for each floating point value $x\in[\alpha,\beta]$ to convert it to a $n$-bit integer $x_q\in[\alpha_q,\beta_q]$:
\begin{equation*}
\scalemath{0.9}{x_q = \left[\frac{1}{\sigma}x+x_0\right],
\sigma=\frac{\beta-\alpha}{\beta_q-\alpha_q}, 
x_0=\left[\frac{\beta\alpha_q-\alpha\beta_q}{\beta-\alpha}\right]
}
\end{equation*}

\todo[disable]{SE: need to reduce font to avoid violating the margin; DL: fixed}
Dynamic quantization~\citep{DBLP:journals/corr/abs-2103-13630} \dl{is a technique that generates the zero-point $x_0$ and scale $\sigma$ parameters in real-time, thereby eliminating the need for calibration data. Due to the unknown distribution parameters, activations are maintained in floating-point format. The process of obtaining quantization parameters $(\alpha,\beta)$ and quantizing floating-point tensors to integer tensors is relatively straightforward, with the necessary statistics being computed during inference.}

Among data-free quantization methods, LLM.int8()~\citep{dettmers2022llmint8} and QLoRA~\citep{dettmers2023qlora} stand out as the most prominent. (i) LLM.int8() quantizes model weights to 8-bit precision using the absmax quantization technique. This method also dynamically quantizes activations to enable efficient matrix multiplications primarily in \textit{int8}, with certain calculations performed in \textit{fp16} for precision. (ii) QLoRA uses a more advanced double quantization approach. It utilizes the \textit{nf4} data type for storage, minimizing memory demands, while computation is conducted in higher precision types (\textit{fp16, bf16}), dequantizing weights on a per-layer basis.

GPTQ~\citep{DBLP:conf/iclr/FrantarAHA23} is an example of weight-only quantization methods. 
It performs layer-by-layer quantization, minimizing the squared error relative to the full precision layer output:
\begin{equation*}
    \label{eq:gptq_objective}
    \argmin_{\widehat{W}} \|WX - \widehat{W} X\|_{F}^2
\end{equation*}
Here\senew{,} $W$ are the full precision weights, $X$ denotes the layer input corresponding to a small set of $m$ data points running through the network, $\widehat{W}$ represents a matrix of quantized weights, and $\|\cdot\|_{F}$ is the Frobenius norm.

\paragraph{Pruning}
\phantomsection
\label{sec:method-pruning}
Pruning is the removal of the least significant parts of the neural network. It can be divided into structured and unstructured. The latter proves helpful on a CPU but is rarely practical on a GPU, since GPUs are heavily optimized for dense matrix multiplication.
Structured pruning can take many forms, from enforcing 2:4 sparsity patterns (in each contiguous block of four values, two values must be zero) to pruning channels or entire blocks of the networks.

Inspired by recent works on layer pruning in LLMs 
\citep{gromov2024unreasonable, men2024shortgpt} which remove 25-50\% of layers with moderate quality drop, we test \se{its applicability for inducing efficient metrics.} 
Specifically, we adopt a simple pruning technique, described in Sec.~4.4 of \citet{gromov2024unreasonable}: in \se{an} $L$-layer model, we drop layers $(L-n)$ to $(L-1)$. This heuristic is based on the observations that pruning deeper layers should affect the model less, as fewer layers rely on changes made by this layer, but also that the ultimate layer is especially important as it ``decodes'' the hidden states for the last fragment of the network, and cannot be removed.
To mitigate the quality drop incurred by layer removal, we apply parameter-efficient fine-tuning.
Concretely, we fine-tune all biases in linear layers, LayerNorm affine parameters, layerwise attention weights, and the regression and tagging heads of xCOMET.
This is akin to \se{the} BitFit~\citep{zaken2022bitfit} sparse-fine-tuning approach, and  has the benefit of adding no parameters and being extremely simple to implement.

We also evaluate magnitude pruning and Wanda pruning \citep{sun2024simple}. In magnitude pruning, the importance of each weight $S_{ij}$ is directly estimated by its magnitude $|W_{ij}|$. Wanda pruning refines this approach by weighting each $|W_{ij}|$ by the average L2 norm of its corresponding input features, $\frac{1}{N} \sum_{j=1}^{N} \|x_j\|_2$, aiming to provide a more informed measure of importance. In both methods, the weights with the lowest importance scores are pruned according to the specified sparsity pattern (unstructured, 2:4 or 4:8).

\paragraph{Distillation}
In distillation, we distinguish between white-box and black-box methods. White-box distillation, detailed in~\citet{DBLP:conf/nips/LiJ22} and \citet{gu2023minillm}, necessitates access to the teacher model internal states, including logits and, possibly, attention maps. This method requires substantial memory and computational resources, as both teacher and student models must be loaded simultaneously, which can be impractical for very large teacher models.

Conversely, black-box distillation, as explored in~\citet{jiang-etal-2023-lion, wu-etal-2024-lamini, DBLP:conf/icml/FuPOSK23}, only requires the teacher model outputs, making it more scalable and feasible for large models or restricted access scenarios. Despite using less information from the teacher, black-box distillation effectively produces high-quality models with reduced computational demands.

For our study, we chose black-box distillation using xCOMET-XXL. This choice allows us to use a very large teacher model, xCOMET-XXL, without encountering the hardware limitations that would arise from white-box distillation. The approach involves using the teacher model to generate pseudo-labels for a large dataset of text triplets. Specifically, the teacher model assigns segment-level quality scores, $q \in [0,1]$, and token-level error span annotations, $k_j \in \{\textit{critical}, \textit{major}, \textit{minor}, \textit{no-error}\}$, for each token in the machine translations, based on MQM annotation guidelines~\citep{freitag-etal-2021-experts}. We simplify the training approach proposed in the original xCOMET paper, adopting a single-phase training method that efficiently trains the student model using these pseudo-labels with both segment-level and word-level supervision. 

\dln{Our approach resembles the recently proposed \textit{Distilling step-by-step} method~\citep{hsieh-etal-2023-distilling}. Both methods utilize black-box distillation without access to the teacher model's internal states. Furthermore, both approaches train the student model on an additional supervision signal beyond the single task-specific label/score. In the case of \textit{Distilling step-by-step}, it is LLM-produced rationales, while in our case, it is error span annotations produced by xCOMET-XXL.}

\section{Experiments}
We compare quantization, pruning, and distillation for compressing xCOMET. We compare it to both released versions, -XL and -XXL. As we focus on computational efficiency, we measure the model \se{\textbf{(i)}} inference speed, \se{\textbf{(ii)}} resource requirements (in terms of GPU memory, vRAM), and \se{\textbf{(iii)}} metric prediction quality, expressed in Kendall-$\tau$ correlation with human judgment. 

\subsection{Evaluation}
\label{sec:04-eval-mqm}
\paragraph{WMT MQM Human Evaluation dataset.} This dataset contains all MQM human annotations from previous WMT Metrics shared tasks~\citep{freitag-etal-2022-results, freitag-etal-2021-results} and from~\citet{freitag-etal-2021-experts}. It contains over 150\se{k} examples for three translation directions (Chinese-English, English-German, English-Russian), five domains (news, TED talks, conversational, social, e-commerce), and three years (2020, 2021, 2022). Following xCOMET~\citep{DBLP:journals/corr/abs-2310-10482}, we use the news 2022 subset (over 16\se{k} samples) for evaluation and the rest of the data for training. 
 
\paragraph{Eval4NLP.} \se{We additionally} use \se{MT} data from \se{the} Eval4NLP shared task~\citep{leiter-etal-2023-eval4nlp}. There are three translation directions: English-Spanish, English-German, and English-Chinese, over 4400 examples in total. No reference translation is provided, which allows to test xCOMET in a reference-free regime.
 
\paragraph{Metric quality evaluation.} We use the Kendall correlation to evaluate the quality of the compared metrics. See Appendix~\ref{appendix:kendall-corr} for \se{a} definition. Each experiment that involves model training is conducted 3 times with different random seeds to account for any fluctuations. We report correlation values obtained by averaging across 3 runs.

\paragraph{Efficiency evaluation.} To evaluate the computational efficiency of compressed models, we measure inference speed in samples per second (samples/s). For a given language pair, we divide the amount of examples by the total time needed to inference the model on the set. Due to the GPU execution and memory models, some operations, such as matrix multiplication, take the same time to execute regardless of the amount of data supplied. \se{Thus,} 
using the largest possible batch size that fits into the accelerator memory is most efficient. To select the optimal batch size, we start with batch size $1$ and increase it by a factor of $2$ until we reach the memory limit on the given GPU.
We test model throughput on RTX 3090 and A100 to explore performance on consumer- and production-level GPUs. Additionally, we provide peak vRAM usage for each model on a fixed batch size \se{of} 8.

\subsection{Setup}
\label{sec:exp-setup}
\paragraph{Quantization.} We use \se{the}  GPTQ~\citep{DBLP:conf/iclr/FrantarAHA23} quantization algorithm and quantize xCOMET to 8, 4, 3, and 2 bits per parameter. We keep default hyperparameters, except using a small subsample of the WikiText2 \citep{merity2016pointer} dataset for calibration. In addition to that, we experiment with data-free quantization methods: LLM.int8() -- 8 bit and QLoRA -- 4 bit. We use \senew{the} implementation from \senew{the} \textit{bitsandbytes} python library.\senew{Initial experiments indicated that}  
models worked faster with their 4-bit quantization implementation if weights were converted to mixed precision before\senew{hand}. This observation was also true for 8-bit quantization, but in this case the quality drop became substantial. Thus, we report LLM.int8() without any uncompressed model transformations, and QLoRA with half-precision model weight conversion.

\paragraph{Pruning.} Following the approach described in the \S\ref{sec:method-pruning}, we apply layer pruning to the underlying encoder model of xCOMET. We remove \se{the} underlying layers from $L-n$ to $L-1$, with $n$ being 4, 8, 12, 16 or 20 layers. We also patch the layerwise attention component of the xCOMET model to reflect changes in the model structure. Subsequently, after pruning, we perform parameter-efficient fine-tuning on the training part of the WMT22 MQM dataset. Fine-tuning is performed for 1 epoch, using AdamW~\citep{DBLP:conf/iclr/LoshchilovH19} optimizer with a learning rate of $1e-4$, effective batch size of 128, and cosine learning rate warmup for 10\% of the duration of training.
\par With Wanda pruning we try 2:4 and 4:8 patterns, to explore setups which can realistically provide speedups on GPU. We use 256 calibration samples from WikiText2\footnote{In \citep{sun2024simple} authors use 128 calibration samples from C4, but we couldn't reproduce the code related to sampling examples from C4.}, and do not finetune pruned model, as the original method does not require it.
We also run simple magnitude pruning with 2:4 and 4:8 sparsity patterns.

\paragraph{Constructing dataset for distillation.} To create a dataset for model compression through distillation, we collected a large number of examples for evaluating MT systems. The collection process involved three main stages. 
\par First, we sampled 500k examples of high-quality parallel texts (source texts and their translations) from the NLLB dataset~\citep{DBLP:journals/corr/abs-2207-04672} for each of the following language pairs: Russian-English, German-English, and Chinese-English. As the NLLB dataset is automatically collected at scale using a bi-text mining model, some translations may be of subpar quality. To address this issue, we applied the xCOMET-XXL model in reference-free mode to filter out examples with low quality scores, which are more likely to be incorrect translations. The filtering threshold was set to the 95th percentile of scores for each language pair, resulting in a threshold of 1.0 (on a 0 to 1 scale) for Russian-English and German-English, and 0.85 for Chinese-English.
\par In the second stage, we generated translation hypotheses for the filtered examples using various MT models with different sizes, architectures, and release dates to ensure high variability in translation quality, following the approach of~\citet{rei-etal-2022-searching}. Additionally, we applied synthetic corruption algorithms to generate hypotheses by corrupting reference translations, as suggested by~\citet{DBLP:journals/corr/abs-2401-17099}. The complete list of models and algorithms used can be found in Appendix~\ref{appendix:data-models}.
\par Finally, in the third stage, we used the xCOMET-XXL model in reference-based mode to generate labels for the collected dataset, including sentence-level scores and error spans. After deduplication and inverting language pairs, our final dataset consists of 14M examples, each containing a source text, reference translation, hypotheses, segment-level quality score and annotated error spans.

\paragraph{Distillation.} We use mDeBERTa v3~\citep{he2023debertav3} as a student. 
It has 278 M parameters~--- 13 times fewer than xCOMET-XL, 39 times fewer than xCOMET-XXL, and 2 times fewer than COMET-22~--- one of the top performers in WMT22 Metrics Shared Task. This model was chosen as it shows superior quality on multilingual language understanding tasks such as XNLI~\citep{conneau-etal-2018-xnli}, compared to alternatives of similar size: InfoXLM~\citep{chi-etal-2021-infoxlm} and XLM-RoBERTa~\citep{conneau-etal-2020-unsupervised}. We trained for 1 epoch, with learning rate of $2e-5$ for scoring head and $1e-5$ for encoder. We set the batch size to 64. Scoring head was configured with two hidden fully connected layers with sizes 3072 and 1024. We compare the prediction quality of the distilled model with original models xCOMET-XL/XXL, as well as with best-performing models of similar size: BLEURT-20~\citep{sellam-etal-2020-bleurt} with 579 M parameters and COMET-22~\citep{rei-etal-2022-comet} with 581 M parameters.

\subsection{Results}

\begin{table*}[tp]
    \centering
    \setlength{\belowcaptionskip}{-10pt}
    \resizebox{0.90\textwidth}{!}{
        \setlength\extrarowheight{-3pt}
        \begin{tabular}{lcccc}
        \toprule
        Model & Compression method & Average Kendall correlation & \makecell{Peak memory consumption (GB) \\ mean (max)} \\
        \cmidrule(lr){1-4}
        XL & None & $0.421$ & 7.76 \; (8.17) \\
        XL & GPTQ 8 bit & $\underline{0.420}$ & 5.20 \; (5.60) \\
        XL & GPTQ 3 bit & $0.408$ & 3.54 \; (3.84) \\
        XL & LLM.int8() & $0.416$ & 7.50 \; (8.32) \\
        XL & QLoRA 4 bit & $0.405$ & 3.75 \; (4.16) \\
        XL & Prune 8 layers & $0.389$ & 6.34 \; (6.66) \\
        XL & Prune 16 layers & $0.365$ & 4.90 \; (5.14) \\
        XL & Magnitude pruning 4:8 & $0.390$ & *7.77 \; (8.18)\\
        XL & Wanda pruning 4:8 & $0.389$ & *8.09 \; (8.25)\\
        \cmidrule(lr){1-4}
        XXL & None & $0.433$ & 22.27 \; (22.39) \\
        XXL & GPTQ 8 bit & $\underline{0.433}$ & 13.81 \; (14.66) \\
        XXL & GPTQ 3 bit & $\underline{0.435}$ & 7.99 \; (8.85) \\
        XXL & LLM.int8() & $0.428$ & 17.86 \; (19.59) \\
        XXL & QLoRA 4 bit & $0.429$ & 9.09 \; (9.94) \\
        XXL & Prune 8 layers & $0.417$ & 19.39 \; (20.09) \\
        XXL & Prune 16 layers & $0.398$ & 15.91 \; (16.48)\\
        XXL & Magnitude pruning 4:8 & $0.418$ & *22.82 \; (23.65) \\
        XXL & Wanda pruning 4:8 & $0.408$ & *22.88 \; (23.65) \\
        \cmidrule(lr){1-4}
        XXL & Distilled~(xCOMET-lite) & $0.388$ & 1.59 \; (1.79) \\
        \bottomrule
        \end{tabular}
    }
    \caption{An overview table with quality / peak memory consumption tradeoff for various representative compression methods in setting \textbf{with} reference translations. Average Kendall correlation and mean/max memory consumption is computed over three language pairs. \underline{Underlined} values indicate compression methods with best prediction quality. XL stands for xCOMET-XL, XXL \senew{stands for} xCOMET-XXL. For Wanda pruning, VRAM consumption is reported using the official method implementation, which stores pruned weights as zeros in original precision. However, potentially 4:8 pruning could deliver almost x2 memory usage reduction.}
    \label{tab:model_performance}
\end{table*}

\begin{table}[h!]
    \centering
    \setlength{\belowcaptionskip}{-15pt}
    \resizebox{\columnwidth}{!}{%
    \begin{tabular}{lccccc}
    \toprule
        Metric & zh-en & en-ru & en-de & Avg. & 
        \se{\#} parameters \\
        \cmidrule(lr){1-6}
        xCOMET-XL &  $0.399$ & $0.414$ & $0.448$ & $0.420$ & $3.5$ B \\
        xCOMET-XXL & $0.390$ & $0.435$ & $0.470$ & $0.432$ & $10.7$ B \\
        \cmidrule(lr){1-6}
        BLEURT-20 & $0.336$ & $0.380$ & $0.379$ & $0.365$ &  $579$ M \\
        COMET-22 & $0.335$ & $0.369$ & \textbf{$0.391$} & $0.361$ & $581$ M \\
        COMETINHO & $0.262$ & $0.330$ & $0.342$ & $0.311$ & $117$ M \\
        xCOMET-lite~(WMT22 data only)
        & $0.280$ & $0.320$ & $0.295$ & $0.298$ & $278$ M \\
        xCOMET-lite & \textbf{$0.360$} & \textbf{$0.422$} & $0.384$ & \textbf{$0.388$} & $278$ M \\
    \bottomrule
    \end{tabular}
    }%
    \caption{Distillation results on WMT MQM News 2022 subset. The numbers are Kendall correlation with human judgement. We compare against BLEURT-20 and COMET-22, which were strong contenders in WMT22 Metrics Shared Task. Additionally\se{,} we compare against a baseline of our model trained on smaller human-annotated dataset WMT22. For reference, there are also scores for large xCOMET models.
    }
    \label{tab:distillation}
\end{table}

\begin{table*}[tp]
    \centering
    \setlength\extrarowheight{-5pt}
    \setlength{\belowcaptionskip}{-15pt}
    \resizebox{0.90\textwidth}{!}{
    \begin{tabular}{llccc}
    \toprule
    Model & Compression method & \makecell{Average \\ Kendall correlation} & \makecell{Samples per second RTX 3090 \\ (min / median / max)} & \makecell{Samples per second A100 \\ (min / median / max) } \\
    \cmidrule(lr){1-5}
    XL & None & $0.421$ & $23.1 \,/\, 30.5 \,/\, 30.9$ & $46.3 \,/\, 59.5 \,/\, 61.8$ \\
    XL & GPTQ 8 bit & $0.420 $ & $10.8 \,/\, 13.7 \,/\, 13.9$ & $29.8 \,/\, 38.5 \,/\, 40.6$ \\
    XL & GPTQ 3 bit & $0.408 $ & $9.9 \,/\, 12.4 \,/\, 12.6$ & $29.6 \,/\, 39.4 \,/\, 40.7$ \\
    XL & LLM.int8() & $0.416 $ & $20.9 \,/\, 28.1 \,/\, 28.5$ & $29.8 \,/\, 38.5 \,/\, 40.6$ \\
    XL & QLoRA 4 bit & $0.405 $ & $22.1 \,/\, 28.8 \,/\, 29.4$ & $44.8 \,/\, 62.9 \,/\, 63.4$ \\
    XL & Prune 8 layers & $0.389 $ & $29.3 \,/\, 38.3 \,/\, 39.1$ & $59.8 \,/\, 72.7 \,/\, 78.5$ \\
    XL & Prune 16 layers & $0.365 $ & $38.6 \,/\, 50.3 \,/\, 51.6$ & $72.0 \,/\, 91.6 \,/\, 96.6$ \\
    XL & Wanda 4:8 & $0.389$ & $25.2 \,/\, 32.8 \,/\, 34.2$ & $56.0 \,/\, 72.1 \,/\, 75.5$ \\
    XL & Magnitude pruning 4:8 & $0.390$ & $25.4 \,/\, 33.4 \,/\, 33.8$ & $53.3 \,/\, 71.7 \,/\, 72.1$ \\
    \cmidrule(lr){1-5}
    XXL & None & $0.433 $ & $7.8 \,/\, 10.0 \,/\, 10.1$ & $17.5 \,/\, 22.5 \,/\, 23.3$ \\
    XXL & GPTQ 8 bit & $0.433 $ & $2.6 \,/\, 3.0 \,/\, 3.0$ & $9.3 \,/\, 11.7 \,/\, 11.9$ \\
    XXL & GPTQ 3 bit & $0.435 $ & $2.7 \,/\, 3.2 \,/\, 3.2$ & $9.0 \,/\, 11.2 \,/\, 11.4$ \\
    XXL & LLM.int8() & $0.428 $ & $9.7 \,/\, 12.4 \,/\, 12.4$ & $13.3 \,/\, 19.0 \,/\, 19.8$ \\
    XXL & QLoRA 4 bit & $0.429 $ & $7.3 \,/\, 9.4 \,/\, 9.5$ & $17.2 \,/\, 22.3 \,/\, 23.3$ \\
    XXL & Prune 8 layers & $0.417 $ & $9.4 \,/\, 12.2 \,/\, 12.3$ & $21.3 \,/\, 26.8 \,/\, 27.6$ \\
    XXL & Prune 16 layers & $0.398 $ & $15.2 \,/\, 15.3 \,/\, 15.5$ & $26.2 \,/\, 33.3 \,/\, 34.3$ \\
    XXL & Wanda pruning 4:8 & $0.408$ & OOM & $ 23.5 \,/\, 29.5 \,/\, 30.5 $ \\
    XXL & Magnitude pruning 4:8 & $0.418$ & OOM & $23.0 \,/\, 29.4 \,/\, 29.6$ \\
    \cmidrule(lr){1-5}
    XXL & Distilled~(xCOMET-lite) & $0.388 $ & $121.4 \,/\, 146.1 \,/\, 153.8$ & $150.5 \,/\, 180.2 \,/\, 190.0$ \\
    \bottomrule
    \end{tabular}
    }
    \caption{Speed results for various methods in settings \textbf{with reference}. Importantly, here the memory consumption is higher than in Table \ref{tab:model_performance}, as we aim for maximal throughput on a given GPU.
    Average Kendall correlation is computed over three language pairs. Samples per second are reported for both 3090 and A100 GPUs. XL stands for xCOMET-XL, XXL stands for xCOMET-XXL. OOM means Out Of Memory error.}
    \label{tab:method_speed_with_reference}
\end{table*}

We present the results of our experiments on quantization, pruning, and distillation. Tables~\ref{tab:model_performance} and \ref{tab:method_speed_with_reference} show the effects of these techniques on xCOMET-XL and xCOMET-XXL models. \dln{Table~\ref{tab:model_performance} focuses on the trade-offs between model quality and memory consumption for pruning and quantization, and Table~\ref{tab:method_speed_with_reference} presents the relationship between model quality and throughput for the same techniques. 
Separately, we present prediction quality for our distilled model in Table~\ref{tab:distillation} and compare it to several baseline metrics of similar size.}

\paragraph{Quantization.} Quantization proves highly effective in reducing \senew{\textbf{memory consumption}} while maintaining quality.
For xCOMET-XL, GPTQ 8-bit achieves nearly identical quality to the baseline, with an average Kendall correlation of 0.420, while reducing peak memory usage by 33\%. 
GPTQ 3-bit provides the largest memory reduction of 54\% at the cost of a 0.013 decrease in correlation.
Notably, xCOMET-XXL sees no quality degradation with GPTQ 8-bit and 3-bit, despite memory reductions of 38\% and 64\%, respectively.
LLM.int8() and QLoRa are suboptimal in terms of quality / peak memory consumption tradeoff, dominated by GPTQ 8-bit and GPTQ 3-bit respectvely.

\par However, as we see in Table~\ref{tab:method_speed_with_reference}, GPTQ slows models down, most likely due to usage of non-optimized CUDA kernels, while QLoRa maintains the thoughput on par with non-compressed model.

\paragraph{Pruning.}
\par Layer pruning substantially improves \textbf{throughput}, particularly for xCOMET-XL. \dln{As we can see in Table~\ref{tab:method_speed_with_reference},} pruning 16 layers provides ~67\% speedup compared to the uncompressed model on an RTX 3090. However, the quality drop is larger compared to quantization methods.
\par Interestingly, magnitude pruning slightly outperforms Wanda pruning, though the latter uses more involved weight importance estimation. Moreover, magnitude pruning performs on par with removing 8 layers, despite keeping only 50\% of non-zero weights. Due to some inefficiencies in official implementation, Wanda pruning and magnitude pruning get OOM error on RTX 3090 on some of the datasets; however, we expect they would show speedups similar to ones on A100. 

\paragraph{Distillation.} Distilling xCOMET-XXL into the much smaller xCOMET-lite model \senew{is} a highly effective compression strategy. \dln{As we demonstrate in Table~\ref{tab:distillation},} despite having only 2.6\% of the parameters (278M vs.\ 10.7B), the distilled model achieves an average Kendall correlation of 0.388, surpassing BLEURT-20 \senew{\&} COMET-22. On English-Russian translation, it even surpasses xCOMET-XL. The effectiveness of using our large-scale \senew{distillation} dataset is further highlighted by the 10-point lower correlation achieved by a model trained on a smaller human-annotated dataset.

\par The distilled xCOMET-lite model offers unparalleled \textbf{speed and memory} efficiency, processing up to 153.8 samples/s on an RTX 3090, 15.2 times faster than the original model (7.8-10.1), \dln{ as we demonstrate in Table~\ref{tab:method_speed_with_reference}. \senew{The} distilled model} has a peak memory consumption of just 1.79 GB, 12.5 times smaller than the original model (22.39 GB). 

\par Additional experiments on reference-free evaluation (Appendix~\ref{sec:appendix}) demonstrate that our distilled model remains competitive with the xCOMET models, achieving an average Kendall correlation of 0.363, just slightly lower than xCOMET-XXL (0.385) and xCOMET-XL (0.378).

\paragraph{Extended Results.} In Appendix~\ref{appendix:detailed-with-ref}, Figure~\ref{fig:wmt-quantization-pruning-colors}, we present detailed results covering all evaluated configurations of pruning and quantization. Notably, 3-bit GPTQ compression maintains prediction quality, contrary to observations in \citet{dettmers2023thecase}, where 4 bits are Pareto-optimal. This suggests that encoder models may be less susceptible to the ``outlier features'' mentioned in \citet{dettmers2022llmint8}. Layer pruning shows promising results for xCOMET-XXL on 4 out of 6 translation directions, with up to 25\% of layers pruned with minimal impact on quality, especially in the reference-free setting.

\subsection{Interaction Analysis}
To further understand the limits of compression of learned metrics for MT evaluation, we \se{explore} interactions between compression methods.

\par We can apply pruning to our distilled model xCOMET-lite to further shrink its size. Given that the encoder now only has 12 layers instead of 48, we \se{evaluate} 3 configurations, pruning 2, 4, or 6 layers from the model. In those experiments, we \se{use} the same hyperparameters as in \S\ref{sec:exp-setup}. We notice a fatal drop in correlation with human judgment by at least 30\% across configurations, to an average score of 0.2645. Please see Table~\ref{tab:interaction-pruning} in Appendix~\ref{appendix:interaction-pruning} for the full results.

We can also apply quantization to the distilled model. Unfortunately, due to architectural details, GPTQ quantization is incompatible with the mDeBERTa architecture. Instead, we apply LLM.int8() and QLoRA quantization (8-bit and 4-bit, respectively). \dln{When comparing the 8-bit quantized xCOMET-lite model to the non-quantized one, we observe only a marginal drop in correlation with human judgment. The 8-bit model achieves an average score of 0.369 across language pairs with references, compared to original xCOMET-lite 0.388. For pairs without references, the 8-bit model scores 0.354, while xCOMET-lite achieves 0.363.}  Notably, the model quantized into 4-bit mode yields a slightly higher correlation for pairs with references\senew{, namely} 0.379. \dln{Furthermore, quantization substantially reduces memory usage. The 8-bit quantization decreases the peak memory consumption of the distilled model by 17\% from 1.8 GB to 1.5 GB, while the 4-bit quantization further reduces it to 1.4 GB.} These results demonstrate that quantization is a viable option for further compressing the distilled model without substantial quality degradation. See Table~\ref{tab:interaction-quant} in Appendix~\ref{appendix:interaction-quant} for full results.

\section{Discussion}
\par The compression methods applied to xCOMET-XL and xCOMET-XXL models demonstrate the potential for reducing memory consumption and increasing processing speed while maintaining competitive prediction quality. Quantization methods, particularly GPTQ 8-bit and 3-bit, achieve substantial memory savings without compromising the models quality. Quantization can also be combined with distillation with little-to-no quality reduction.

\par Pruning methods, while capable of reducing memory consumption and increasing throughput, result in a more noticeable decrease in correlation compared to quantization. Our results align with the findings in~\citet{rei-etal-2022-searching}, which conclude that up to 5 out of 24 layers of encoder model can be removed without noticeable quality degradation of the metric. At the same time, the layer pruning works slightly worse than in other tasks \citep{gromov2024unreasonable, men2024shortgpt}, where up to 50\% of layers could be removed for large models. Pruning appears incompatible with our distilled model, due to a substantial drop in \se{metric quality.}
Magnitude pruning with 4:8 sparsity pattern shows promising results with respect to quality / speedup trade-off. Moreover, it potentially offers almost 50\% reduction in peak memory consumption (and e.g. \texttt{torch} library will likely support structured spasity formats quite soon).

\par The distillation of xCOMET-XXL into the smaller mDeBERTa-based model, xCOMET-lite, is a highly effective approach for improving computational efficiency while maintaining competitive metric quality. Our distillation method, based on collecting large-scale diverse  dataset, proves successful for distilling the xCOMET metric and is easily scalable to additional translation directions. 

\par When considering speed, the distilled xCOMET-lite outperforms other compression methods, processing a substantially higher number of samples per second on both consumer-grade RTX 3090 and HPC-grade A100 GPUs. Pruning is the next best performer, allowing for up to 1.3-1.5 times speedup while maintaining competitive metric quality.

\section{Conclusion}

In the rapidly evolving field of MT evaluation, the current top-performing metrics, such as MetricX, xCOMET, and GEMBA-MQM, are all based on extremely large underlying models. These models, including mT5 with 13B parameters, XLM-RoBERTa-XXL with 11B parameters, and the closed-source GPT-4 with an estimated 1.7T parameters, pushed the boundaries of performance but come with \se{substantial} computational costs and hardware requirements.

Our research aims to address these challenges by comparing three commonly used compression methods ~--- quantization, pruning, and knowledge distillation ~--- in compressing the xCOMET model. We have demonstrated that these methods can effectively reduce memory consumption and increase processing speed while maintaining competitive performance, making them viable options for deploying large state-of-the-art learned metric for MT evaluation in a resource-constrained environments. In particular, our distilled model xCOMET-lite achieves competitive prediction quality with a substantially smaller model size, offering a solution for researchers and practitioners with no access to top-tier hardware.

\dl{Based on our findings, we recommend the following: for the highest quality with a reduced VRAM requirements, opt for 8-bit or 3-bit quantization with GPTQ.  For improved speed without substantial quality penalty, test 4-bit quantization with QLoRA, try structured magnitude pruning (2:4, 4:8) or prune up to 25\% of the model layers.For massive speedup and low hardware requirements, consider the distilled model xCOMET-lite or its quantized version, accepting a slight compromise on quality. The choice of compression method ultimately depends on the hardware, amount of data, and acceptable quality loss.}


\section*{Acknowledgments}
\par The NLLG group gratefully acknowledges support from the Federal Ministry of Education and Research (BMBF) via the research grant ``Metrics4NLG'' and the German Research Foundation (DFG) via the Heisenberg Grant EG 375/5-1.

\par The NLLG group acknowledges support by the state of Baden-Württemberg through bwHPC.

\section{Limitations} 
While our research provides valuable insights into the compression of large language models for machine translation evaluation, it is important to acknowledge the limitations of our work. 
\begin{itemize}
\setlength{\itemsep}{0pt}
\setlength{\parskip}{0pt}
\setlength{\parsep}{0pt} 
\item Our study focuses solely on machine translation evaluation and does not consider other tasks, such as summarization evaluation. To the best of our knowledge, all currently existing summarization evaluation metrics are regression-only and do not offer error span prediction. Therefore, it is unclear if the results would be different for this task. Future research could explore the applicability of these compression methods to a broader range of natural language processing tasks.

\item Our measure of a metric quality, Kendall-$\tau$ correlation with human judgments, is known to incorrectly reward metrics for predicting ties~\citep{deutsch-etal-2023-ties}.
\item Although our research has potential implications for low-resource machine translation, we did not conduct experiments on low-resource language pairs. We plan to address this limitation when releasing the subsequent versions of our models to the public. 
\item Our distillation approach still requires the availability of the original teacher model. Training such a model is expensive in terms of both computational resources and the cost of human annotation for the training data.
\end{itemize}

\bibliography{anthology,custom}

\appendix
\section{Models and Algorithms used for Data Collection}
\label{appendix:data-models}
\begin{itemize}
    \item OPUS-MT~\citep{tiedemann-thottingal-2020-opus} monodirectional models: \textit{en-ru}, \textit{ru-en}, \textit{en-zh}, \textit{zh-en}, \textit{en-de}, \textit{de-en}.
    \item OPUS-MT models for multiple languages: \textit{mul-en} and \textit{en-mul}.
    \item NLLB models~\citep{DBLP:journals/corr/abs-2207-04672}, versions: Distilled 600M and 1.3B, Non-Distilled 1.3B and 3.3B.
    \item Word Drop: it was used to create translation hypotheses by randomly dropping 15\% of the words from reference translation.
    \item Word Replacement with MLM: similarly we applied XLM-RoBERTa-Large for masked language modelling task to replace 15\% of the words.
    \item Backtranslation: we applied NLLB-1.3B model to translate references into a proxy language and back. As a proxy languages we used French and Japanese.
    \item Backtranslation + MLM: consists of applying MLM to the results of backtranslation.
\end{itemize}

\section{Kendall Correlation}
\label{appendix:kendall-corr}
Kendall-$\tau$ correlation is defined as follows: let $(x_1, y_1), \ldots, (x_n, y_n)$ 
be observations of random variables $X$ and $Y$ 
such that all values of $x_i$ and $y_i$ are unique. A pair of observations $(x_i, y_i)$ and $(x_j, y_j)$ is said to be \textit{concordant} if either $x_i < x_j; \; y_i < y_j$ or $x_i > x_j;\; y_i > y_j$, otherwise this pair is \textit{discordant}. The Kendall correlation coefficient $\tau$ is \todo[disable]{SE: this description can be relegated to the appendix} 
\begin{align*}
\tau = \frac{n_c - n_d}{C^2_n} = \frac{C^2_n - n_d - n_d}{C^2_n} \\
= 1 - \frac{2 n_d}{C^2_n} = 1 - \frac{4 \cdot n_d}{n(n - 1)} \end{align*}
where $n$ is the total amount of observations, $n_c$ is the amount of concordant pairs, and $n_d$ is the amount of discordant pairs. Kendall correlation coefficient is more robust to outliers than Pearson correlation and better captures non-linear dependencies. In our case, $X$ is the ground truth MQM score, and $Y$ is the score estimated by the neural metric.

\section{Interaction Analysis of Distillation and Pruning}
\label{appendix:interaction-pruning}
\begin{table}[H]
    \centering
    \resizebox{\columnwidth}{!}{%
    \begin{tabular}{ccc}
    \toprule
     \# pruned layers & \makecell{Avg. correlation \\ with ref.} & \makecell{Avg. correlation \\ without ref.} \\
    \cmidrule(lr){1-3}
    2 & 0.240 & 0.209 \\
    4 & 0.264 & 0.202 \\
    6 & 0.201 & 0.181 \\
    \bottomrule
    \end{tabular}
    }%
    \caption{Results of evaluation of xCOMET-lite distilled from xCOMET-XXL with applied pruning. \textit{Avg. correlation} represents Kendall correlation averaged across 3 language pairs.}
    \label{tab:interaction-pruning}
\end{table}

\section{Interaction Analysis of Distillation and Quantization}
\label{appendix:interaction-quant}
\begin{table}[H]
    \centering
    \resizebox{\columnwidth}{!}{%
    \begin{tabular}{ccccc}
    \toprule
     Method & \# bits & \makecell{Avg. correlation \\ with ref.} & \makecell{Avg. correlation \\ without ref.} & Peak Mem. Cons. (GB) \\
    \cmidrule(lr){1-5}
    LLM.int8() & 8 & 0.369 & 0.355 & 1.2 (1.5)\\
    QLoRA & 4 & 0.379 & 0.345 & 1.1 (1.4)\\
    \bottomrule
    \end{tabular}
    }%
    \caption{Results of evaluation of xCOMET-lite distilled from xCOMET-XXL with applied quantization. \textit{Avg. correlation} represents Kendall correlation averaged across 3 language pairs.}
    \label{tab:interaction-quant}
\end{table}

\section{Detailed results on compression and quantization}
\label{appendix:detailed-with-ref}
\begin{figure*}[bp]
    \centering
    \includegraphics[width=\textwidth]{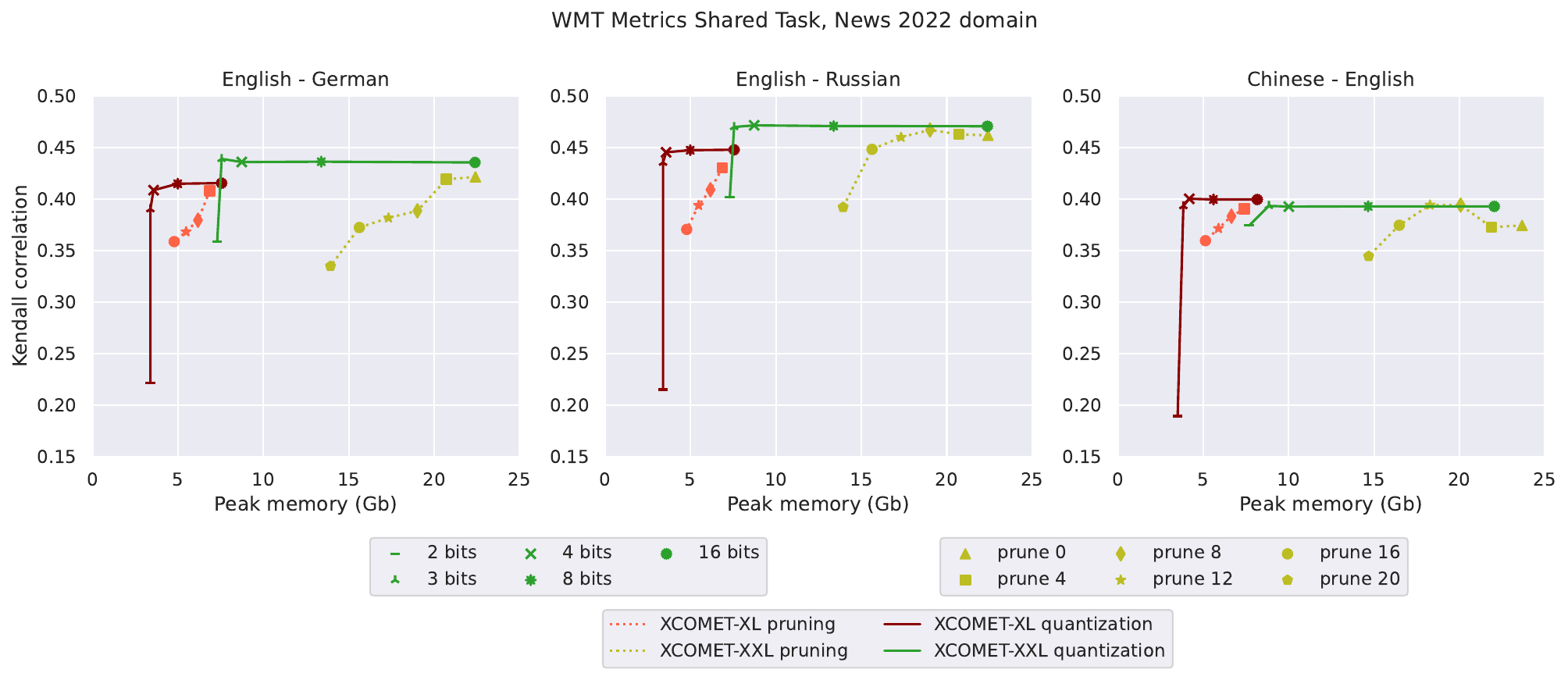}
    \caption{Results on WMT MQM Human Evaluation dataset. In this setting xCOMET has access to reference translation.}
    \label{fig:wmt-quantization-pruning-colors}
\end{figure*}

See Figure~\ref{fig:wmt-quantization-pruning-colors}.

\section{Results on Eval4NLP dataset}
\label{sec:appendix}

\begin{figure*}[p]
   \centering
   \includegraphics[width=\textwidth]{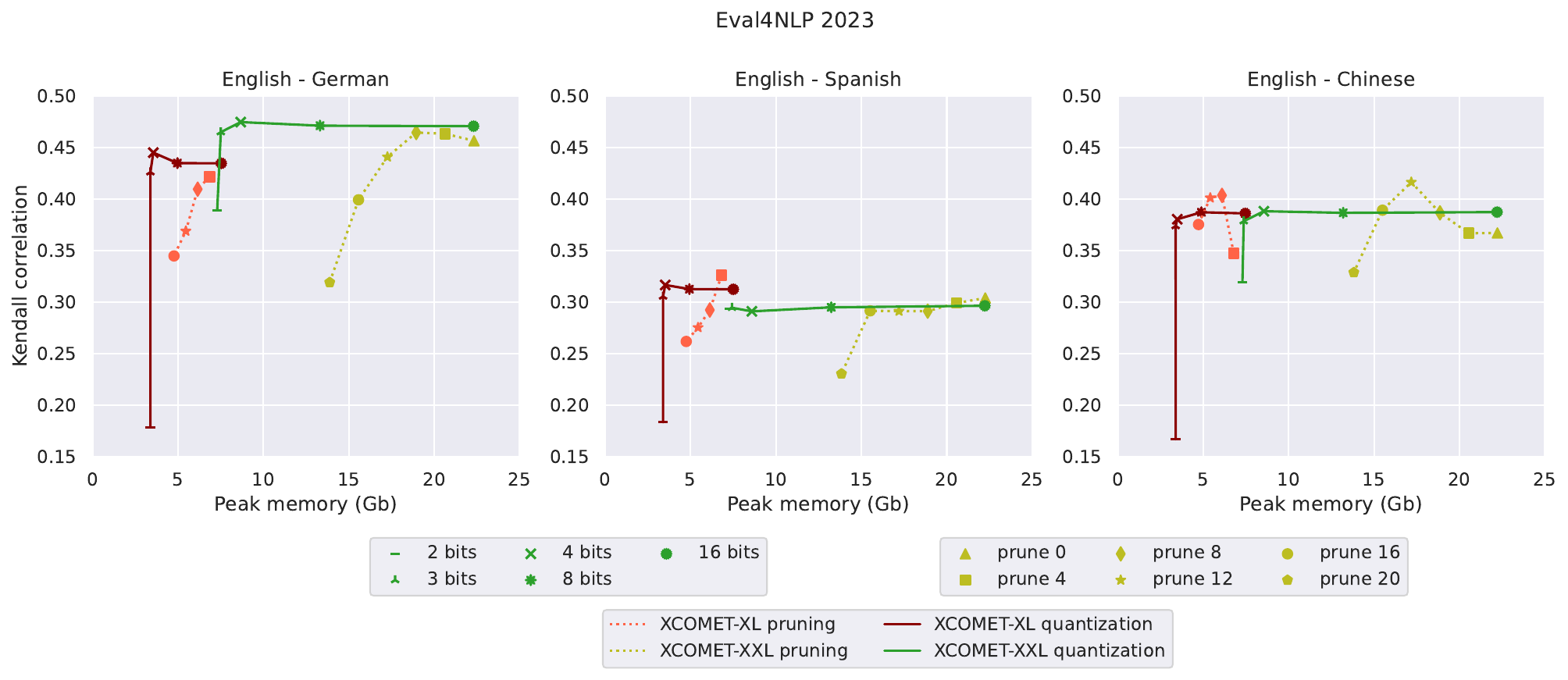}
   \caption{Results on Eval4NLP dataset. This is reference-free setting, also known as Quality Estimation (QE).}
   \label{fig:eval4nlp-quantization-pruning-colors}
\end{figure*}

\begin{table*}[tp]
    \centering
    \resizebox{\textwidth}{!}{%
    \begin{tabular}{lcccc}
    \toprule
    Model & Compression method & Average Kendall correlation & Peak memory consumption (Gb) \\
    \cmidrule(lr){1-4}
    XL & None & $0.378$ & 7.51 \; (7.54) \\
    XL & GPTQ 8 bit & $0.379$ & 4.94 \; (4.97) \\
    XL & GPTQ 3 bit & $0.372$ & 3.39 \; (3.39) \\
    XL & LLM.int8() & $0.384$ & 6.98 \; (7.06) \\
    XL & QLoRA 4 bit & $0.373$ & 3.50 \; (3.53) \\
    XL & Prune 8 layers & $0.373$ & 6.13 \; (6.16) \\
    XL & Prune 16 layers & $0.359$ & 4.75 \; (4.77) \\
    XL & Magnitude pruning 4:8 & $0.362$ & 7.51 \; (7.55) \\
    XL & Wanda pruning 4:8 & $0.342$ & 8.01 \; (8.01) \\
    \cmidrule(lr){1-4}
    XXL & None & $0.385$ & 22.24 \; (22.30) \\
    XXL & GPTQ 8 bit & $0.385$ & 13.25 \; (13.32) \\
    XXL & GPTQ 3 bit & $0.378$ & 7.44 \; (7.51) \\
    XXL & LLM.int8() & $0.383$ & 16.78 \; (16.94) \\
    XXL & QLoRA 4 bit & $0.373$ & 9.09 \; (9.94) \\
    XXL & Prune 8 layers & $0.381$ & 18.91 \; (18.97) \\
    XXL & Prune 16 layers & $0.360$ & 15.53 \; (15.57) \\
    XXL & Magnitude pruning 4:8 & $0.340$ & 22.25 \; (22.31) \\
    XXL & Wanda pruning 4:8 & $0.340$ & 22.50 \; (22.50) \\
    \cmidrule(lr){1-4}
    XXL & Distilled~(xCOMET-lite) & $0.363$ & 1.4 \;(1.4) \\
    \bottomrule
    \end{tabular}
    }
    \caption{An overview table with some representative results for various compression methods in setting \textbf{without} reference translations. Average is computed over three language pairs for Kendall correlation. For peak memory the mean and maximum values are computed, and the maximum is reported in parentheses. XL stands for xCOMET-XL, XXL -- xCOMET-XXL.}
    \label{tab:model_performance_no_reference}
\end{table*}

\begin{table*}[tp]
    \centering
    \resizebox{0.95\textwidth}{!}{
    \begin{tabular}{l l c c c}
    \toprule
    Model & Compression method & \makecell{Average \\ Kendall correlation} & \makecell{Samples per second RTX 3090 \\ (min / median / max)} & \makecell{Samples per second A100 \\ (min / median / max)} \\
    \cmidrule(lr){1-5}
    XL & None & $0.378$ & $55.1 \,/\, 67.0 \,/\, 70.5$ & $76.2 \,/\, 98.9 \,/\, 111.8$ \\
    XL & GPTQ 8 bit & $0.379$ & $30.8 \,/\, 35.1 \,/\, 35.9$ & $53.0 \,/\, 69.0 \,/\, 72.9$ \\
    XL & GPTQ 3 bit & $0.372$ & $28.7 \,/\, 33.3 \,/\, 33.7$ & $57.3 \,/\, 71.3 \,/\, 74.0$ \\
    XL & LLM.int8() & $0.384$ & $50.9 \,/\, 59.7 \,/\, 63.8$ & $64.1 \,/\, 87.2 \,/\, 88.1$ \\
    XL & QLoRA 4 bit & $0.373$ & $55.0 \,/\, 66.2 \,/\, 68.7$ & $93.0 \,/\, 123.2 \,/\, 135.5$ \\
    XL & Prune 8 layers & $0.373$ & $70.5 \,/\, 85.6 \,/\, 87.7$ & $94.5 \,/\, 119.5 \,/\, 131.2$ \\
    XL & Prune 16 layers & $0.359$ & $82.9 \,/\, 108.6 \,/\, 110.2$ & $110.4 \,/\, 128.3 \,/\, 149.2$ \\
    \cmidrule(lr){1-5}
    XXL & None & $0.385$ & $22.1 \,/\, 24.2 \,/\, 25.2$ & $35.4 \,/\, 48.3 \,/\, 48.6$ \\
    XXL & GPTQ 8 bit & $0.385$ & $8.1 \,/\, 8.5 \,/\, 8.5$ & $23.7 \,/\, 28.9 \,/\, 29.6$ \\
    XXL & GPTQ 3 bit & $0.378$ & $8.6 \,/\, 9.2 \,/\, 9.3$ & $20.9 \,/\, 23.9 \,/\, 29.1$ \\
    XXL & LLM.int8() & $0.383$ & $27.8 \,/\, 30.8 \,/\, 32.1$ & $38.3 \,/\, 48.2 \,/\, 48.9$ \\
    XXL & QLoRA 4 bit & $0.373$ & $21.8 \,/\, 25.2 \,/\, 25.5$ & $42.6 \,/\, 51.9 \,/\, 57.4$ \\
    XXL & Prune 8 layers & $0.381$ & $25.4 \,/\, 28.4 \,/\, 29.8$ & $42.6 \,/\, 56.7 \,/\, 60.2$ \\
    XXL & Prune 16 layers & $0.360$ & $30.0 \,/\, 34.8 \,/\, 36.3$ & $50.8 \,/\, 64.4 \,/\, 68.1$ \\
    XXL & Distilled~(xCOMET-lite) & 0.363 & $312.1 \,/\, 352.0 \,/\, 358.0$ & $229.0 \,/\, 232.2 \,/\, 241.9$ \\
    \bottomrule
    \end{tabular}
    }
    \caption{Speed results for different methods in setting \textbf{without reference}. Importantly, here the memory consumption is higher than in Table \ref{tab:model_performance_no_reference}, as we aim for maximal throughput on a given GPU. Average and std are computed over three language pairs for Kendall correlation. Samples per second are reported for both 3090 and A100 GPUs. XL stands for xCOMET-XL, XXL -- xCOMET-XXL.}
    \label{tab:method_speed_no_reference}
\end{table*}

In addition to WMT Shared Metric dataset, we perform evaluations on Eval4NLP dataset, in setting without reference translation. The results are shown on Figure~\ref{fig:eval4nlp-quantization-pruning-colors} and Tables \ref{tab:model_performance_no_reference}, \ref{tab:method_speed_no_reference}. All conclusions are stable with respect to another dataset.

\section{Varying seed for layer pruning}
To check the robustness of finetuning procedure in layer pruning technique, we run the same pipeline with three seeds. The standard deviations are presented in Table \ref{tab:pruning_seed}.

\begin{table*}[tp]
    \centering
    \setlength{\belowcaptionskip}{-10pt}
    \resizebox{\textwidth}{!}{
        \setlength\extrarowheight{-3pt}
        \begin{tabular}{lcccccc}
        \toprule
        Model & Compression method & Chinese - English & English - Russian & English - German & Peak memory consumption (GB)  \\
        \cmidrule(lr){1-6}
        XL & None & $0.399$ & $0.448$ & $0.415$ & 7.76 \; (8.17) \\
        XL & Prune 8 layers & $0.387 \pm 0.005$ & $0.414 \pm 0.006$ & $0.381 \pm 0.004$ & 6.34 \; (6.66) \\
        XL & Prune 16 layers & $0.362 \pm 0.002$ & $0.369 \pm 0.006$ & $0.359 \pm 0.009$ & 4.90 \; (5.14) \\
        \cmidrule(lr){1-6}
        XXL & None & $0.390$ & $0.470$ & $0.435$ & 22.27 \; (22.39) \\
        XXL & Prune 8 layers & $0.398 \pm 0.000$ & $0.435 \pm 0.000$ & $0.385 \pm 0.000$ & 19.39 \; (20.09) \\
        XXL & Prune 16 layers & $0.372 \pm 0.001$ & $0.445 \pm 0.004$ & $0.352 \pm 0.006$ & 15.91 \; (16.48) \\
        \bottomrule
        \end{tabular}
    }
    \caption{Robustness of layer pruning approach to random seed, setting \textbf{with} reference translations. For peak memory consumption, the mean and maximum values are computed, and the maximum is reported in parentheses. XL stands for xCOMET-XL, XXL -- xCOMET-XXL.}
    \label{tab:pruning_seed}
\end{table*}

\section{Additional Details}
In this section we discuss some additional details concerning our research.

\subsection{Risks}
\par While our work demonstrates the potential of distillation, quantization, and pruning techniques in creating an efficient alternative to xCOMET, there are some risks to consider:
\begin{itemize}
\item The use of distilled models like xCOMET-lite, as well as over-pruned models, in high-stakes applications, such as filtering datasets or evaluating machine translation systems in sensitive domains (e.g., healthcare, legal), may lead to suboptimal decisions due to the slightly lower accuracy compared to the full xCOMET model. One must exercise discretion when considering acceptable loss of quality.
\item Our work primarily focuses on high-resource languages, and the performance of the compressed models on low-resource languages remains unexplored. The lack of training data and the potential differences in linguistic characteristics may lead to suboptimal performance when applying these models to evaluate translations in low-resource language pairs. This could result in inaccurate quality assessments and hinder the development of reliable machine translation systems for these languages.
\item The availability of highly efficient evaluation metrics like xCOMET-lite may prompt researchers and practitioners to conduct large-scale experiments, such as web-scale dataset filtration or extensive hyperparameter optimization. While these experiments can lead to valuable insights and improvements in machine translation systems, they may also consume substantial amounts of computational resources and power. This increased energy consumption could contribute to environmental concerns and raise questions about the sustainability of such practices.
\end{itemize}

\subsection{Artifacts}
\par The main artifact that we use in our research is a set of two pre-trained metrics for MT evaluation: xCOMET-XL and xCOMET-XXL, released by~\cite{DBLP:journals/corr/abs-2310-10482}. Those models are released under~\textit{cc-by-nc-sa-4.0} license. Our use of these models complies with the license and is consistent with usage permissions.
\par We plan to release two of our own artifacts: the distilled model xCOMET-lite and the dataset that was used to train it. Both of those artifacts will also be released under~\textit{cc-by-nc-sa-4.0} according to the ``share-alike'' requirement of this license, as derivatives of the original xCOMET models.

\subsection{PII in the dataset}
According to the dataset card of the NLLB dataset~\footnote{https://huggingface.co/datasets/allenai/nllb}, the data may contain personally identifiable information (PII). Identifying and anonymizng such information is outside of the scope of this work. We plan to address it in future, before releasing dataset to the public.

\subsection{Used packages}
In our experiments we use the following key software libraries:
\begin{itemize}
    \item PyTorch: v2.0.1
    \item Transformers: v4.41.2
    \item BitsAndBytes: v0.41.1
    \item AutoGPTQ: v0.7.0
    \item Optimum: v1.11.0
    \item SciPy: v1.11.1
    \item Unbabel COMET: v2.0.2
\end{itemize}

\end{document}